\documentclass[10pt,twocolumn,letterpaper]{article}
\usepackage{iccv}
\usepackage{times}
\usepackage{epsfig}
\usepackage{graphicx}
\usepackage{amsmath}
\usepackage{amssymb}

\usepackage{ragged2e}
\usepackage{booktabs}
\usepackage{color}
\usepackage{multirow}
\usepackage{bm}
\usepackage{bbm}
\usepackage{amsmath}
\usepackage{algorithm}
\usepackage{algorithmic}
\usepackage{amsfonts}
\usepackage{enumerate}

\usepackage{cuted}
\usepackage{capt-of}

\def\ie{{\em i.e.}}
\def\eg{{\em e.g.}}
\def\etal{{\em et al.}}

\newcommand{\figref}[1]{Fig. \ref{#1}}
\newcommand{\tabref}[1]{Tab. \ref{#1}}

\newcommand{\mc}[1]{\mathcal{#1}}

\newcommand{\br}[1]{\bm{\mathrm{#1}}}

\usepackage{amssymb}
\usepackage{pifont}
\newcommand{\cmark}{\ding{51}}%
\newcommand{\xmark}{\ding{55}}%

\usepackage[pagebackref=true,breaklinks=true,letterpaper=true,colorlinks,bookmarks=false]{hyperref}

\iccvfinalcopy 

\def\iccvPaperID{2947} 

\ificcvfinal\fi

\begin{document}

\title{FACIAL: Synthesizing Dynamic Talking Face with Implicit Attribute Learning}

\author{
Chenxu Zhang\textsuperscript{1}, Yifan Zhao\textsuperscript{2}, Yifei Huang\textsuperscript{3}, Ming Zeng\textsuperscript{4}, Saifeng Ni\textsuperscript{5}\\
Madhukar Budagavi\textsuperscript{5}, Xiaohu Guo\textsuperscript{1}\\
\textsuperscript{1}The University of Texas at Dallas\qquad\textsuperscript{2}Beihang University\qquad\textsuperscript{3}East China Normal University\\
\textsuperscript{4}Xiamen University\qquad\textsuperscript{5}Samsung Research America\\
{\tt\small \{chenxu.zhang, xguo\}@utdallas.edu, zhaoyf@buaa.edu.cn, yifeihuang17@gmail.com}\\ {\tt\small zengming@xmu.edu.cn, \{saifeng.ni, m.budagavi\}@samsung.com}

}

\maketitle
\ificcvfinal\thispagestyle{empty}\fi

\begin{strip}\centering
\includegraphics[width=0.9\textwidth]{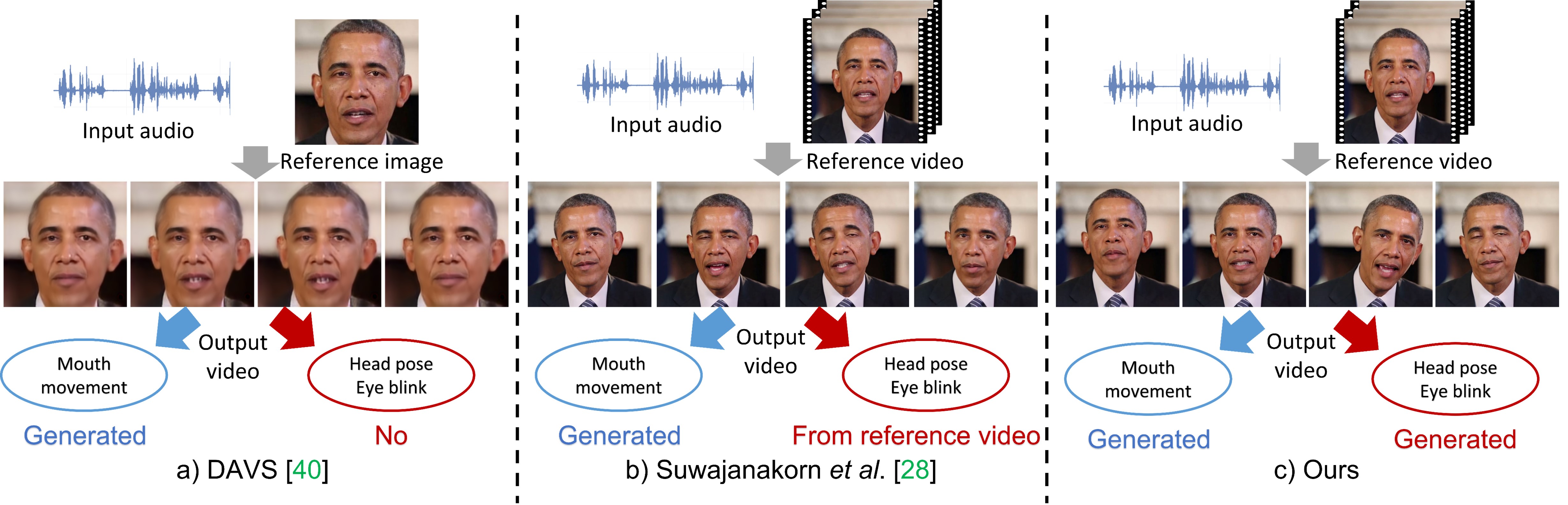}
\captionof{figure}{Illustration of three typical frameworks. a) Explicit attribute generation: only considering mouth movements of talking head. b) Explicit generation with implicit morphing: only generating explicit mouth movements and taking the implicit attributes from reference videos. c) Our implicit attribute learning framework: generating explicit and implicit attributes from input audio in one unified framework.
\label{fig:feature-graphic}}
\end{strip}

\begin{abstract}
In this paper, we propose a talking face generation method that takes an audio signal as input and a short target video clip as reference, and synthesizes a photo-realistic video of the target face with natural lip motions, head poses, and eye blinks that are in-sync with the input audio signal.
We note that the synthetic face attributes include not only explicit ones such as lip motions that have high correlations with speech, but also implicit ones such as head poses and eye blinks that have only weak correlation with the input audio. To model such complicated relationships among different face attributes with input audio, we propose a FACe Implicit Attribute Learning Generative Adversarial Network (FACIAL-GAN), which integrates the phonetics-aware, context-aware, and identity-aware information to synthesize the 3D face animation with realistic motions of lips, head poses, and eye blinks. Then, our Rendering-to-Video network takes the rendered face images and the attention map of eye blinks as input to generate the photo-realistic output video frames. 
Experimental results and user studies show our method can generate realistic talking face videos with not only synchronized lip motions, but also natural head movements and eye blinks, with better qualities than the results of state-of-the-art methods.

\end{abstract}

\section{Introduction}
Synthesizing dynamic talking faces driven by input audio has become an important technique in computer vision, computer graphics, and virtual reality. 
There have been steady research progresses~\cite{bregler1997video,chen2019hierarchical,chung2017b,ezzat2002trainable,sinha2020identity,song2018talking,vougioukas2019end,vougioukas2019realistic,zhou2019talking}, however, it is still very challenging to generate photo-realistic talking faces that are indistinguishable from real captured videos, which not only contain synchronized lip motions, but also have personalized and natural head movements and eye blinks, etc.

The information contained in dynamic talking faces can be roughly categorized into two different levels: 1) the attributes that need to be synchronized with the input audio,~\eg, the lip motion that has strong correlations with the signals of auditory phonetics; 2) the attributes that have only weak correlations with the phonetic signal,~\eg, the head motion that is related to both the context of speech and the personalized talking style and the eye blinking whose rate is mainly decided by personal health condition as well as external stimulus. Here we call the first type of attributes to be \emph{explicit} attributes, and the second type to be \emph{implicit} attributes.

It should be noted that the majority of existing works~\cite{chen2019hierarchical,chung2017b,ezzat2002trainable,sinha2020identity,song2018talking,zhou2019talking} on talking face generation are focusing on \emph{explicit} attributes only, by synchronizing the lip motions with input audio. 
Examples include Zhou~\etal~\cite{zhou2019talking} disentangled the audio into subjected-related information and speech-related information to generate clear lip patterns, and Chen~\etal's Audio Transformation and Visual Generation (ATVG) networks~\cite{chen2019hierarchical} to transfer audio to facial landmarks and generate video frames conditioned on landmarks. There are only a few recent efforts~\cite{chen2020talking,yi2020audio,zhou2020makelttalk} exploring the correlation between the \emph{implicit} attributes of head pose with the input audio. For example, Chen~\etal~\cite{chen2020talking} adopted a multi-layer perceptron as the head pose learner to predict the transformation matrix of each input frame. However, it remains unclear on: (1) how the \emph{explicit} and \emph{implicit} attributes might potentially influence each other? (2) how to model implicit attributes, such as head poses and eye blinks, that depend not only on the \emph{phonetic} signal, but also on the \emph{contextual} information of speech as well as the \emph{personalized} talking style?

\begin{figure*}[t]
\begin{center}
\includegraphics[width=0.98\linewidth]{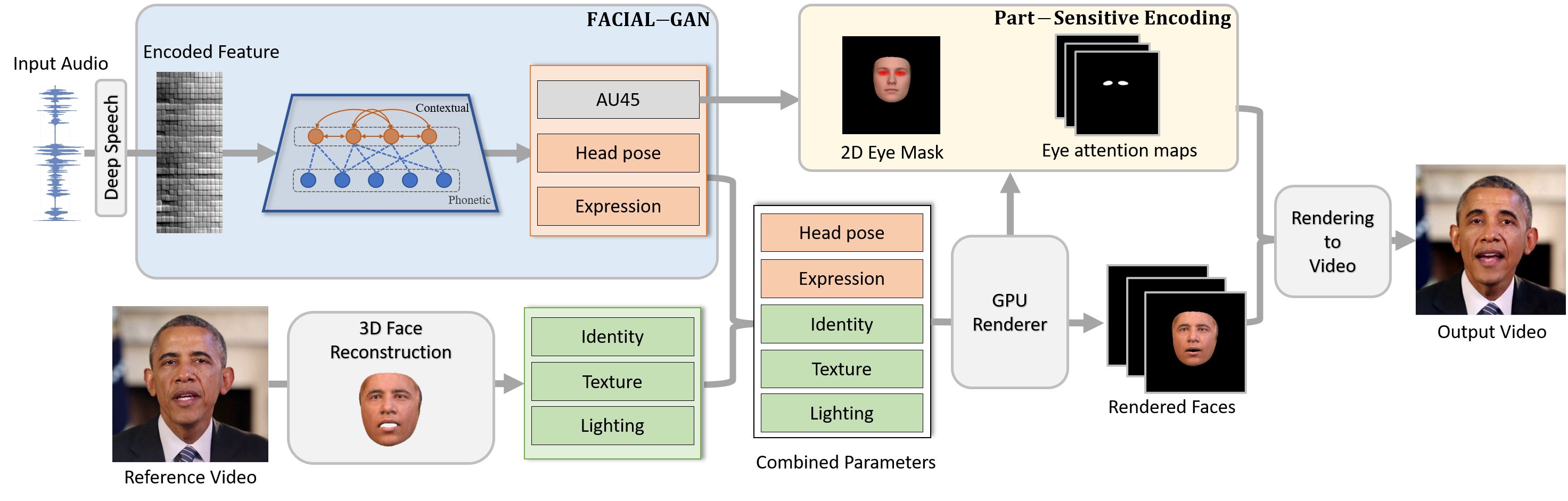}
\end{center}
   \caption{Overview of the proposed implicit attribute learning framework. Given input audio, the proposed FACIAL-GAN aims to generate the explicit attributes (expression) and implicit attributes (eye blinking AU45, head pose) with jointly temporal correlations and local phonetic features. The reference video is performed with a face reconstruction operation to provide a 3D model guidance for rendering operation.
   Besides, a part-sensitive encoding takes the eye blinking action units as input and serves as eye attention maps for rendering faces. These guidances are jointly combined to feed into the rendering-to-video network.
   }
\label{fig:pipeline}
\end{figure*}

To tackle these challenges, we propose a FACe Implicit Attribute Learning (FACIAL) framework for synthesizing dynamic talking faces, as shown in~\figref{fig:pipeline}.
(1) Unlike the previous work~\cite{chen2020talking} predicting implicit attributes using an individual head pose learner, our FACIAL framework jointly learns the \textit{implicit} and \textit{explicit} attributes with the regularization of adversarial learning. We propose to embed all attributes, including Action Unit (AU) of eye blinking, head pose, expression, identity, texture and lighting, in a collaborative manner so their potential interactions for talking face generation can be modeled under the same framework. (2) We design a special FACIAL-GAN in this framework to jointly learn \emph{phonetic}, \emph{contextual}, and \emph{personalized} information. It takes a sequence of frames as a grouped input and generates a contextual latent vector, which is further encoded together with the phonetic information of each frame, by individual frame-based generators. FACIAL-GAN is initially trained on our whole dataset (Sec.~\ref{sec:dataset}). Given a short reference video ($2\sim 3$ minutes) of the target subject, FACIAL-GAN will be fine-tuned with this short video, so it can capture the personalized information contained in it. Hence our FACIAL-GAN can well-capture all phonetic, contextual, and personalized information of the implicit attributes, such as head poses. (3) Our FACIAL-GAN can also predict the AU of eye blinks, which is further embedded into an auxiliary eye-attention map for the final Rendering-to-Video module, to generate realistic eye blinking in the synthesized talking face.

With the joint learning of explicit and implicit attributes, our end-to-end FACIAL framework can generate photo-realistic dynamic talking faces as shown in~\figref{fig:feature-graphic}, superior to the results produced by the state-of-the-art methods. 
The \textbf{contribution} of this paper is threefold: 
(1) We propose a joint explicit and implicit attribute learning framework to synthesize photo-realistic talking face videos with audio-synchronized lip motion, personalized and natural head motion, and realistic eye blinks. 
(2) We design a FACIAL-GAN module to encode the contextual information with the phonetic information of each individual frame, to model the implicit attributes needed for synthesizing natural head motions.
(3) We embed the FACIAL-GAN generated AU of eye blinking into an eye-attention map of rendered faces, which achieves realistic eye blinks in the resulting video produced by the Rendering-to-Video module.

\section{Related Work}
\textbf{Audio-driven talking face generation}
Most of existing talking face generation methods~\cite{bregler1997video,chen2019hierarchical,chung2017b,ezzat2002trainable,sinha2020identity,song2018talking,vougioukas2019end,vougioukas2019realistic,zhou2019talking} focus on generating videos that are in sync with the input audio stream. Chung \textit{et al.}~\cite{chung2017b} proposed an encoder-decoder CNN model using a joint embedding of face and audio to generate synthesized talking face video frames. Chen \textit{et al.}~\cite{chen2019hierarchical} proposed a hierarchical structure to first transfer audio to landmarks and then to generate video frames conditioned on landmarks. However, in talking face videos generated by these approaches, the head pose is almost fixed during the speech. To achieve photo-realistic videos with head motion, several techniques~\cite{suwajanakorn2017synthesizing,prajwal2020lip,thies2020neural,wen2020photorealistic} first generate the lip area that is in sync with input audio and compose it into an original video. Suwajanakorn \textit{et al.}~\cite{suwajanakorn2017synthesizing} used an audio stream from Barack Obama to synthesize a photo-realistic video of his speech. However, this method is applicable to other characters because of the requirement of a large amount of video footage. Thies \textit{et al.}~\cite{thies2020neural} employed a latent 3D model space to generate talking face video which can be used for different people. However, those methods cannot disentangle the head motion and facial expression due to their intrinsic limitations, which means the head motion is irrelevant with the input audio. More recently, Chen \textit{et al.}~\cite{chen2020talking} and Yi \textit{et al.}~\cite{yi2020audio} focus on generating head movement directly from input audio. Yi \textit{et al.}~\cite{yi2020audio} proposed a memory-augmented GAN module to generate photo-realistic videos with personalized head poses. However, due to the limitations of network and 3D model, their generated face expression (\eg, eye blinks) and head motion tend to be still. In comparison, we introduce the FACIAL-GAN module to integrate phonetic, contextual, and personalized information of the talking, and combine the synthesized 3D model with AU attention map to generate photo-realistic videos with synchronized lip motion, personalized and natural head poses and eye blinks.

\textbf{Video-driven talking face generation}
Video-driven talking face generation methods~\cite{kim2018deep,kim2019neural,zhang2019one,zakharov2019few,pumarola2018ganimation,wiles2018x2face,thies2016face2face} transferred face expression and slight head movements from the given source video frames to target video frames.
Zakharov \textit{et al.}~\cite{zakharov2019few} presented a system to frame the few- and one-shot learning of neural talking head models of unseen people as adversarial training problems with high capacity generators and discriminators. 
Kim \textit{et al.}~\cite{kim2018deep} introduced a generative neural network to transfer head pose, face expression, eye gaze and blinks from a source actor to a portrait video based on the generated 3D model. However, since the head movements and face expressions are guided by the source video, these methods can only generate pre-determined talking head movements and expressions which is consistent with the source ones. 

\section{Approach}

\subsection{Problem Formulation}
Given an input audio $\mc{A}$ and a short ($2\sim 3$ minutes) reference video $\mc{V}$ of the subject, our talking head synthesis aims to generate a speech video $\mc{S}$ of the subject synchronized with $\mc{A}$. The conventional steps to generate neural talking head can be represented as:
\begin{equation}
\begin{split}
\mc{F}_{lip} &= \br{G}(\br{E}(\mc{A})), \\
\mc{S} &= \br{R} (\mc{F}_{lip},\mc{V}),
\end{split}
\end{equation}
where $\mc{F}_{lip}$ denotes the explicit features synthesized by an adversarial generator $\br{G}$. $\br{E}$ denotes audio feature extraction network and $\br{R}$ denotes the rendering network to translate the synthesized features into the output video.  

As mentioned above, this conventional synthesizing approach usually fails to capture the implicit attributes,~\eg, dynamic head poses $\mc{M}_{pose}$, and eye blinks $\mc{M}_{eye}$. Toward this end, we further exploit the intrinsic interrelationships among speech audio and these implicit attributes, namely FACe Implicit Attribute Learning (FACIAL). Besides, we introduce an auxiliary part attention map of eye regions $\mc{E}$. Our FACIAL synthesis process has the form:
\begin{equation}
\begin{split}
\{ \mc{F}_{lip},\mc{M}_{pose},\mc{M}_{eye} \}= \br{G}(\br{E}(\mc{A})),\\ 
\mc{S} = \br{R} (\mc{F}_{lip},\mc{M}_{pose},\mc{E} \odot \mc{M}_{eye},\mc{V}).
\end{split}
\end{equation}
The overall framework in~\figref{fig:pipeline} is composed of two essential parts,~\ie, a FACIAL Generative Adversarial Network (FACIAL-GAN) to encode the joint explicit and implicit attributes, and a Rendering-to-Video network to synthesize the output talking face video with synchronized lip motion, natural head pose, and realistic eye blinks. Furthermore, different attributes require individual encoding strategies, the explicit attributes $\mc{F}_{lip}$ are highly correlated with the syllables of input audio, which are decided by each audio frame. However, implicit features $\mc{M}_ { \{ eye, pose \}}$ heavily relies on the long-term information,~\eg, head movements of the next frames are decided by the previous states. We thus elaborate on how to embed these attributes into one unified framework in the next subsections.

\subsection{FACIAL-GAN}
To jointly embed the explicit and implicit attributes in one unified network, we need to: 1) generate explicit expressions corresponding to phonetic features of each frame; 2) embed the contextual information,~\ie, temporal correlations into the network for implicit attribute learning. We propose FACIAL-GAN as a solution to achieve these goals.

\begin{figure}
\begin{center}
\includegraphics[width=1\columnwidth]{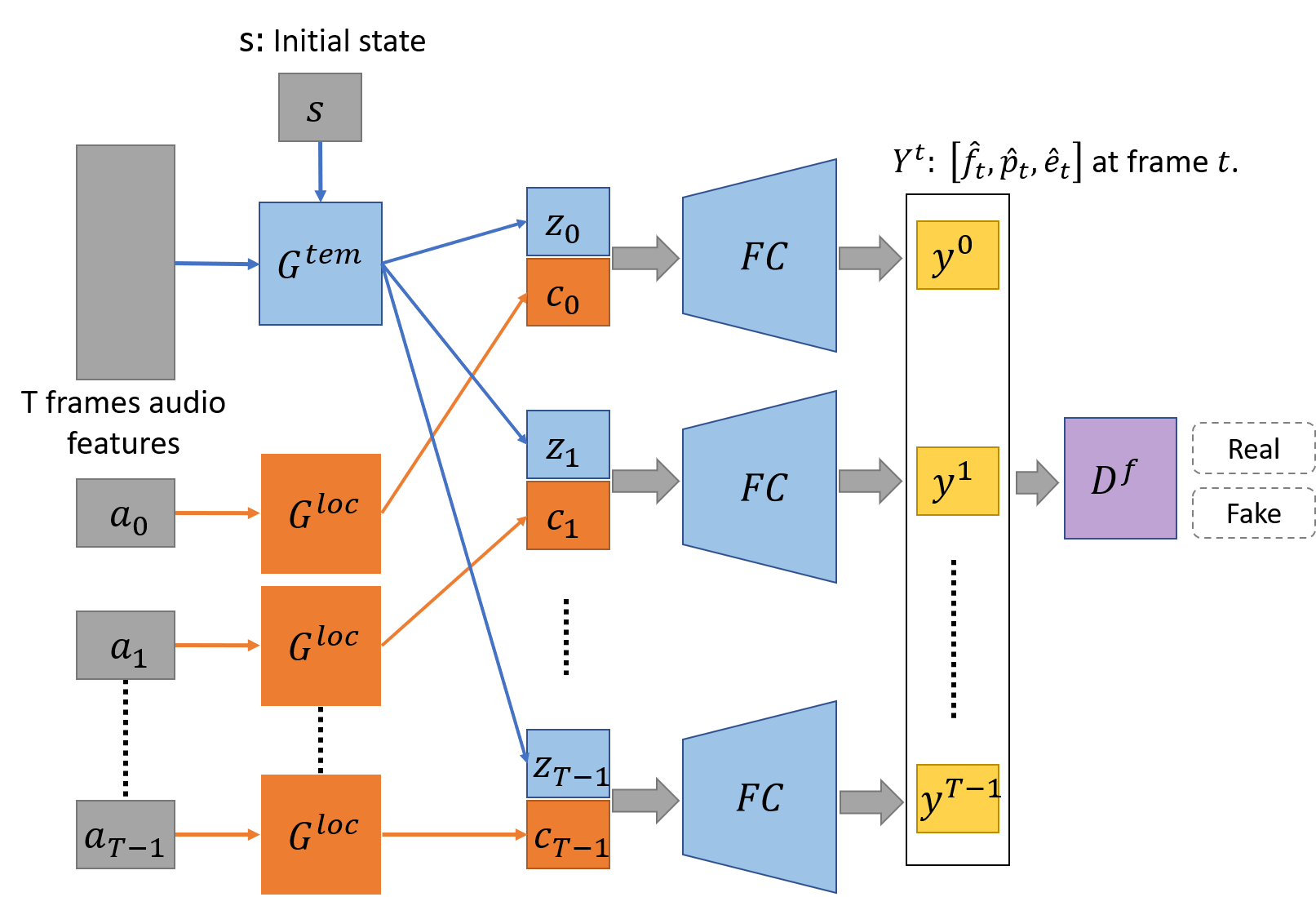}
\end{center}
  \caption{Framework of the proposed FACIAL-GAN. $\br{G}^{tem}$ takes the whole sequence of $T$ frames as input to generate temporal vector $z$, while $\br{G}^{loc}$ generates the local latent vector $c$ of each frame.  }
\label{fig:facialgan}
\vspace{-0.3cm}
\end{figure}

The proposed FACIAL-GAN is composed of three essential parts: temporal correlation generator $\br{G}^{tem}$ to build contextual relationships and a local phonetic generator $\br{G}^{loc}$ to extract the characteristics of each frame. Besides, a discriminator network $\br{D}^{f}$ is employed to judge real or fake of the generated attributes.
As shown in~\figref{fig:facialgan}, the input audio $\mc{A}$ is sampled by a sliding window of $T$ frames, and is prepossessed with DeepSpeech~\cite{hannun2014deep} to generate the feature $\br{a} \in \mathbb{R}^{29 \times T}$. Let $\br{f}$ denote the facial expression parameters, $\br{p}$ denote head pose features and $\br{e}$ denote eye blink AU estimation, and use $f_t$, $p_t$ and $e_t$ to represent the features of the $t$-th frame, respectively (See supplementary for details). 

\textbf{Temporal correlation generator:}
To extract the temporal correlations of the whole input sequence, our key idea to feed the audio sequence $\mc{A}$ of $T$ frames into a contextual encoder, which generates the latent global features $\br{z}$. 
As taking the audio sequence as a whole, each unit of latent feature $\br{z}$ is able to incorporate the information from other frames. Hence the corresponding feature $z_t$ of the $t$-th frame can be extracted by splitting the encoded feature $\br{z}$.
Given the DeepSpeech features $a[0:T-1]$ of input audio $\mc{A}$ and its initial state $\br{s} = \{ f_0,p_0,e_0 \} \in \mathbb{R}^{71}$, we generate the predicted temporal attribute sequence ${z}_t$, $t\in[0, T-1]$ with $\br{G}^{tem}$. The initial state $\br{s}$ is introduced to ensure the temporal continuity between generated sequences.

\textbf{Local phonetic generator:}
The temporal network $\br{G}^{tem}$ focuses on the whole temporal domain, where the phonetic features of each frame are not emphasized. Therefore, we employ the local phonetic network $\br{G}^{loc}$ to generate local features $c_t$ for the $t$-th frame. Taking the $t$-th frame as an example, $\br{G}^{loc}$ takes audio features $a_t=a[t-8:t+8]$ as input and outputs the local feature $c_t$. Now, we have obtained the temporal features $z_t$ and local features $c_t$ for time step $t$. 
One fully connected layer $\br{FC}$ is used to map $z_t$ and $c_t$ to the predicted parameters ${ \hat{f}_t,\hat{p}_t,\hat{e}_t } \in \mathbb{R}^{71}$. 
The encoding process of FACIAL-GAN can be represented as:
\begin{equation}
\begin{split}
& z_t=\mc{S}(\br{G}^{tem}(\br{E}(\mc{A})|\br{s}),t),
\\
& c_t = \br{G}^{loc}(\mc{S}(\br{E}(\mc{A}),t)),
\\
& [\hat{f}_t,\hat{p}_t,\hat{e}_t]=\br{FC}(z_t \oplus c_t),
\end{split}
\end{equation}
where function $\mc{S}(\br{X},t)$ denotes splitting and extracting the $t$-th feature block of feature $\br{X}$,  and $\oplus$ is the feature concatenation operation. $\br{E}$ denotes the audio feature extraction.

\textbf{Learning objective:}
We supervise our generator networks $\br{G}^{tem}$ and $\br{G}^{loc}$ with the following loss functions:
\begin{equation} \label{lossreg}
\mathcal{L}_{\text{Reg}} = \omega_1\mathcal{L}_{\text{exp}} + \omega_2 \mathcal{L}_{\text{pose}} + \omega_3 \mathcal{L}_{\text{eye}} + \omega_4  \mathcal{L}_{s},
\end{equation}
where $\omega_1$, $\omega_2$, $\omega_3$, and $\omega_4$ are the balancing weights, and $\mathcal{L}_{s}$ is the $L_1$ norm loss for the initial state values, which guarantees the continuity between the generated sequences of sliding windows:
\begin{equation} 
\mathcal{L}_{s} = \|f_0-\hat{f}_0\|_1+\|p_0-\hat{p}_0\|_1+\|e_0-\hat{e}_0\|_1.
\end{equation}
$\mathcal{L}_{\text{exp}}$, $\mathcal{L}_{\text{pose}}$,  and $\mathcal{L}_{\text{eye}}$ are the $L_2$ norm losses for facial expression, head pose, and eye blink AU respectively. We also introduce the motion loss $\mc{U}$ to guarantee the inter-frame continuity:
\begin{equation} \label{lossl2}
\begin{split}
\mathcal{L}_{exp} &= \sum_{t=0}^{T-1}\mc{V}(f_t,\hat{f}_t)+\omega_5\sum_{t=1}^{T-1}\mc{U}(f_{t-1},f_t,\hat{f}_{t-1},\hat{f}_t),
\\
\mathcal{L}_{\text{pose}} &= \sum_{t=0}^{T-1}\mc{V}(p_t,\hat{p}_t)+\omega_5\sum_{t=1}^{T-1}\mc{U}(p_{t-1},p_t,\hat{p}_{t-1},\hat{p}_t),
\\
\mathcal{L}_{\text{eye}} &= \sum_{t=0}^{T-1}\mc{V}(e_t,\hat{e}_t)+\omega_5\sum_{t=1}^{T-1}\mc{U}(e_{t-1},e_t,\hat{e}_{t-1},\hat{e}_t),
\end{split}
\end{equation}
where $\mc{V}(x_t,\hat{x}_t)=||x_t-\hat{x}_t||^2_2$, and $\mc{U}(x_{t-1},x_t,\hat{x}_{t-1},\hat{x}_{t})= \|x_t-x_{t-1}-(\hat{x}_{t}-\hat{x}_{t-1})\|^2_2$ is to guarantee the temporal consistency between adjacent frames. $\omega_5$ is a weight to balance these two terms. Here we use $x$ to represent the ground-truth values of the predicted $\hat{x}$.

The loss of facial discriminator $\br{D}^{f}$ is defined as:
\begin{equation} \label{discriminatorfun}
\begin{split}
\mathcal{L}_{\text{F-GAN}} = 
& \arg \min_{\br{G}^f} \max_{\br{D}^{f}} \mathbb{E}_{\mathbf{f},\mathbf{p},\mathbf{e}}[\log \br{D}^f(\mathbf{f},\mathbf{p},\mathbf{e})]+\\
& \qquad \mathbb{E}_{\mathbf{a},\mathbf{s}}[\log (1 - \br{D}^f(\br{G}^f(\mathbf{a},\mathbf{s}))],
\end{split}
\end{equation}
where the generator $\br{G}^f$ is composed of two sub-generators $\br{G}^{tem}$ and $\br{G}^{loc}$, minimizing this objective function, while discriminator $\br{D}^f$ are optimized for maximization.
The final loss function is then defined as:
\begin{equation} \label{lossall}
\begin{split}
& \mathcal{L}_{facial}= \omega_6 \mathcal{L}_{\text{F-GAN}} + \mathcal{L}_{\text{Reg}}.
\end{split}
\end{equation}

\begin{figure}[t]
\begin{center}
\includegraphics[width=0.98\linewidth]{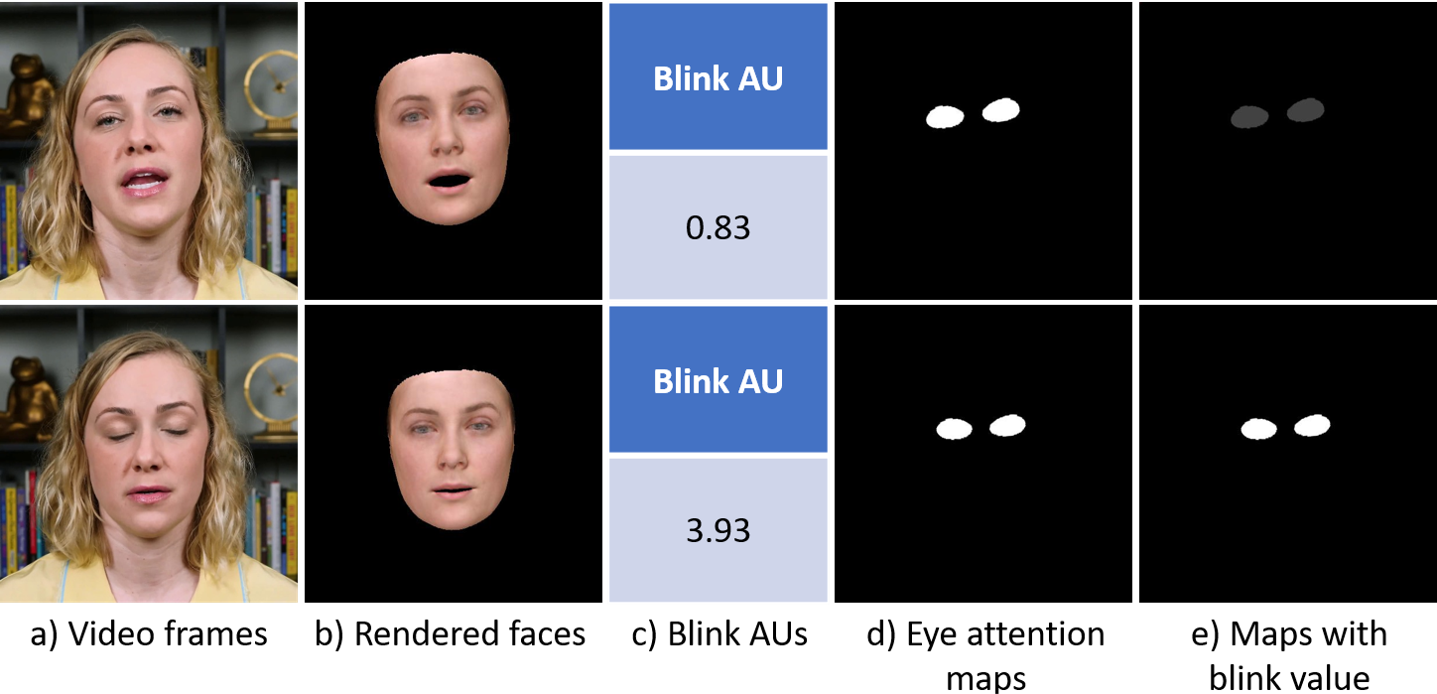}
\end{center}
\caption{Illustration of the part sensitive encoding map. Our final generated encoding e) is composed of two parts: c) estimated blink AU, and d) eye attention maps. }
\label{fig:eyeattention}
\end{figure}

\subsection{Implicit Part-Sensitive Encoding}
With the combination of geometry, texture and illumination coefficients from reference video and generated expression and head pose coefficients from input audio, we can render the 3D face with personalized head movements. 3D model describes the head pose better than 2D methods by rotating and translating the head. However, it is very difficult for 3D reconstruction methods to capture the subtle motions in the upper face region, especially for eye blinks as shown in~\figref{fig:eyeattention}. We combine the advantages of both 3D model and 2D action units to generate talking face with personalized head movements and natural eye blinks.

An intuitive solution is to directly concatenate the blink value to face image channels. However, convolutional neural networks can not recognize this channel for the eye part. We propose to use an eye attention map, which first locates the eye region, and then only changes the pixel value of this region according to the blinking AU value. 

We first mark the vertices of the eye regions in 3D models.
The vertices are identified from the mean face geometry of 3D Morphable Model (3DMM) by the following criteria:
\begin{equation}
(v_x-center_x)^2/4+(v_y-center_y)^2<th,
\end{equation}
where $v_x,v_y$ are the x, y values of vertex $v$, and $center_x$, $center_y$ are the x, y values of the center of  each eye landmark. Threshold $th$ is used to adjust the size of eye regions.

During the 3D face rendering, we locate pixels related with marked regions to generate the eye attention map for each face image in~\figref{fig:eyeattention}. Finally, we apply the normalized blinking value to the pixels in the eye attention map.

\subsection{Rendering-to-Video Network}
We employ the rendering-to-video network to translate the rendering images into the final photo-realistic images. Inspired by Kim \textit{et al.}~\cite{kim2018deep}, we first combine the rendering image with the eye attention map to generate the training input data $\hat{I}$ with a size of $W\times H \times 4$ (rendering image with 3 channels, attention map with 1 channel). To ensure temporal coherency, we use a window of size $2N_w$ with the current frame at the center of the window. 

By following Chan \textit{et al.}'s method~\cite{chan2019everybody}, we train our rendering-to-video network consisting of a generator $\br{G}^{r}$ and a multi-scale discriminator $\br{D}^{r} = (\br{D}^{r}_1,\br{D}^{r}_2,\br{D}^{r}_3)$ that are optimized alternatively in an adversarial manner. The generator $\br{G}^{r}$ takes the stacked tensor $X_t=\{\hat{I}_t\}_{t-N_w}^{t+N_w}$ of size $W\times H \times 8N_w$ as input and outputs a photo-realistic image $\br{G}^{r}(X_t)$ of the target person. The conditional discriminator $\br{D}^{r}$ takes the stacked tensor $X_t$ and a checking frame (either a real image $I$ or a generated image $\br{G}^{r}(X_t)$) as input and discriminates whether the checking frame is real or not. 
The loss function can be formulated as:
\begin{equation} \label{lossgd}
\begin{split}
\mathcal{L}_{render} = & \sum_{\br{D}_i^{r} \in \br{D}^{r}} (\mathcal{L}_{R-GAN}(\br{G}^{r},\br{D}^{r}_i) + \lambda_1\mathcal{L}_{FM}(\br{G}^{r},\br{D}^{r}_i)) \\
&+ \lambda_2\mathcal{L}_{VGG}(\br{G}^{r}(X_t),I) + \lambda_3\mathcal{L}_{1}(\br{G}^{r}(X_t),I),
\end{split}
\end{equation}
where $\mathcal{L}_{R-GAN}(\br{G}^{r},\br{D}^{r})$ is the GAN adversarial loss, $\mathcal{L}_{FM}(\br{G}^{r},\br{D}^{r})$ denotes the discriminator feature-matching loss proposed by~\cite{wang2018high}, $\mathcal{L}_{VGG}(\br{G}^{r},I)$ is a VGG perceptual loss~\cite{johnson2016perceptual} for semantic level similarities, and $\mathcal{L}_{1}(\br{G}^{r},I)$ is an absolute pixel error loss. 

The optimal network parameters can be obtained by solving a typical min-max optimization:
\begin{equation}
\br{G}^{r*} = \arg \min_{\br{G}^{r}} \max_{\br{D}^{r}}\mathcal{L}_{render}(\br{G}^{r},\br{D}^{r}).
\end{equation}

\section{Dataset Collection}\label{sec:dataset}

As mentioned above, previous popular datasets mostly neglect the combination of explicit and implicit attributes. For example, GRID~\cite{cooke2006audio} provides a fixed head pose for talking head video, and some other datasets do not focus on the attributes of one specific person,~\eg, LRW~\cite{chung2016lip} includes many short clips of different people. 
To jointly incorporate the explicit and implicit attributes for neural talking heads, we adopt the talking head dataset from Zhang \textit{et al.}~\cite{zhang2021face}, with rich information,~\ie, dynamic head poses, eye motions, lip synchronization as well as the 3D face model for each frame in an automatic collection manner.

\begin{figure*}
\begin{center}
\includegraphics[width=0.93\linewidth]{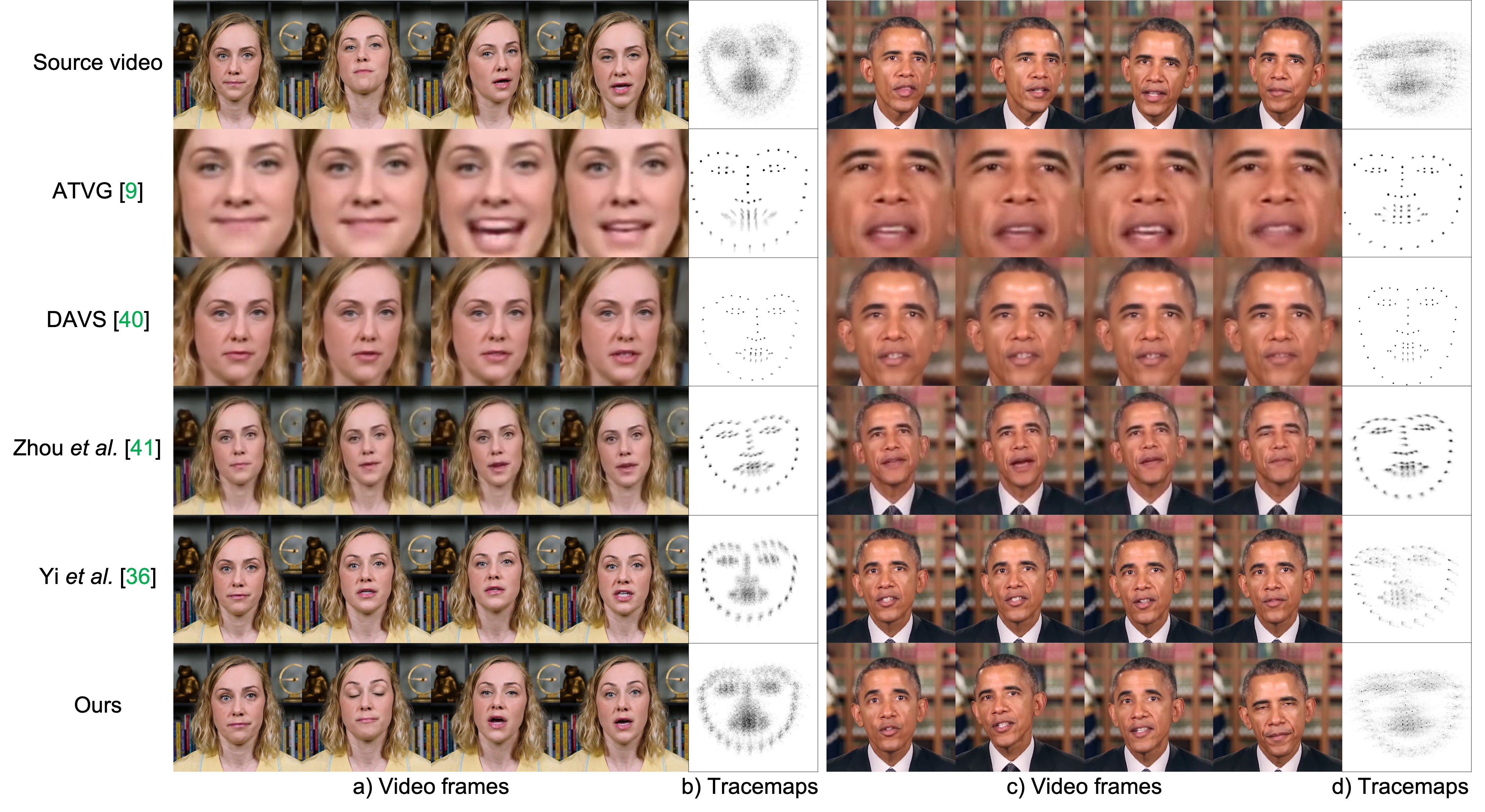}
\end{center}
 \caption{Comparisons with ATVG~\cite{chen2019hierarchical}, DAVS~\cite{zhou2019talking}, Zhou~\etal~\cite{zhou2020makelttalk} and Yi~\etal~\cite{yi2020audio}. The first row is corresponding video frames with input audio. a) and c) are the generated video frames. b) and d) are the corresponding tracemaps of facial landmarks in multiple frames. From the tracemaps we can see our generated head motions are highly consistent with the source videos.}
\label{fig:comparison1}
\end{figure*}

\textbf{Audio preprocessing.}
We employ DeepSpeech~\cite{hannun2014deep} to extract speech features. DeepSpeech outputs the normalized log probabilities of characters in 50 Frames Per Second (FPS), which forms an array of size $50\times D$ for each second. Here $D = 29$ is the number of speech features in each frame. We resample the output to 30 FPS using linear interpolation to match the video frames in our dataset, which generates an array of size $30\times D$ for each second.

\textbf{Head pose and eye motion field.}
To automatically collect the head pose as well as to detect the eye motions, we adopt  OpenFace~\cite{amos2016openface} for the face parameter generation of each video frame.
The rigid head pose $\mathbf{p}\in \mathbb{R}^6$ is represented by Euler angles (pitch $\theta_x$, yaw $\theta_y$, roll $\theta_z$) and a 3D translation vector $\mathbf{t} \in \mathbb{R}^3$. To depict the eye motions, Action Units (AUs)~\cite{ekman1978manual} are exploited to define the action intensities of muscle groups around the eye regions.

\textbf{3D face reconstruction.}
To automatically generate 3D face models, we adopt Deng \textit{et al.}'s method~\cite{deng2019accurate} to generate face parameters [$F_\text{id}, F_\text{exp}$,$F_\text{tex}$,$\gamma$], where $F_\text{id} \in \mathbb{R}^{80}$, $F_\text{exp} \in \mathbb{R}^{64}$ and $F_\text{tex} \in \mathbb{R}^{80}$ are the coefficients for geometry, expression, and texture, respectively for the 3D Morphable Model (3DMM)~\cite{blanz1999morphable}. $\gamma \in \mathbb{R}^{27}$ is the spherical harmonics (SH)~\cite{ramamoorthi2001efficient} illumination coefficients.
The parametric face model of 3DMM consists of a template triangle mesh with $N$ vertices and an affine model which defines face geometry $S\in \mathbb{R}^{3N}$ and texture $T\in \mathbb{R}^{3N}$: 
\begin{equation}
\begin{split}
& S = \overline{S} + B_\text{id}F_\text{id} + B_\text{exp}F_\text{exp},
\\
& T = \overline{T} + B_\text{tex}F_\text{tex},
\end{split}
\end{equation}
where $\overline{S}$ and $\overline{T}\in \mathbb{R}^{3N}$ denote the averaged face geometry and texture, respectively. $B_\text{id}$, $B_\text{tex}$ and $B_\text{exp}$ are the PCA basis of geometry,texture and expression adopted from the Basel Face Model~\cite{paysan20093d} and FaceWareHouse~\cite{cao2013facewarehouse}.

\textbf{Dataset statistics.}
The proposed dataset contains rich samples of more than 450 video clips which are collected from the videos used by Agarwal~\textit{et al.}~\cite{agarwal2019protecting}. Each video clip lasts for around $1$ minute. We re-normalize all videos to 30 FPS, forming 535,400 frames in total. We further divide our dataset using a train-val-test split of 5-1-4. Each video clip in our dataset has a stable fixed camera and appropriate lighting, with only one speaker for stable face generation.

\section{Experiments}

\subsection{Network Learning}

\textbf{Training.} Our training scheme consists of two steps: (1) We first optimize FACIAL-GAN loss $\mc{L}_{facial}$ based on our whole training dataset, which mainly considers the general mapping between the audio and the generated attributes. (2) Given a reference video $\mc{V}$, we first extract the audio feature $a$, 3D face model, head pose $p$ and eye blinking AU $e$. Then we fine-tune the FACIAL-GAN loss $\mc{L}_{facial}$ to learn the personalized talking style. Meanwhile, we optimize the rendering loss $\mc{L}_{render}$ to learn the mapping from rendered faces with eye attention maps to the final video frame.
    
\textbf{Testing.} Given the input audio, we first use the fine-tuned FACIAL-GAN to map the audio features to expression $\br{f}$, head pose $\br{p}$ and eye blink AU $\br{e}$, which have the personalized talking style of the reference video. Then, we render the corresponding face images and eye attention maps, and convert them to a photo-realistic target video with the personalized talking style.

\textbf{Implementation details.}
All experiments are conducted on a single NVIDIA 1080-Ti GPU using Adam~\cite{kingma2014adam} optimizer and a learning rate of 0.0001.

We use a sliding window of size $T=128$ to extract training samples of audio and video, and use a sliding distance of 5 frames between neighboring samples. A total of 50 epochs are trained with a batch size of 64 for general training. For the fine-tuning step, it takes 10 epochs with a batch size of 16.
For the rendering-to-video network, the training process takes 50 epochs with a batch size of 1 with a learning rate decay for the last 30 epochs. 
In our experiments, the parameters in Eqs.~\eqref{lossreg}, \eqref{lossl2}, \eqref{lossall} are $\omega_1 = 2$, $\omega_2 = 1$, $\omega_3 = 5$, $\omega_4 = 10$, $\omega_5 = 10$, and $\omega_6 = 0.1$. The parameters in Eq.~\eqref{lossgd} are $\lambda_1 = 2$, $\lambda_2 = 10$, and $\lambda_3 = 50$.

\begin{figure}[t]
\begin{center}
\includegraphics[width=0.98\linewidth]{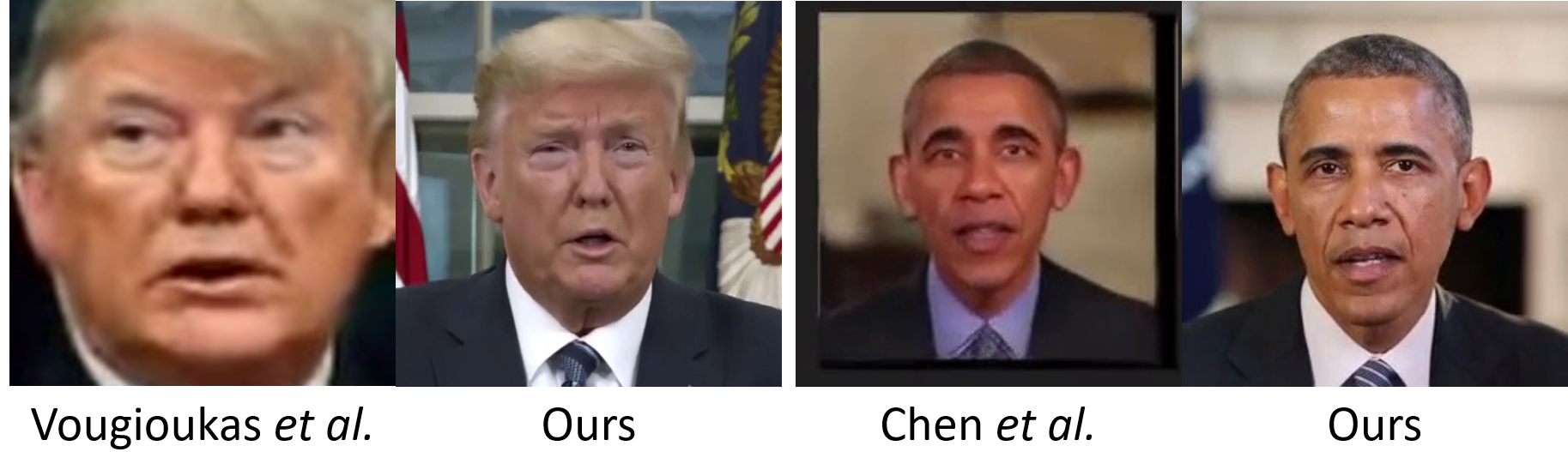}
\end{center}
\caption{Comparison to 2D GAN-based Vougioukas \textit{et al.}'s~\cite{vougioukas2019realistic} and Chen \textit{et al.}'s~\cite{chen2020talking} methods.}
\label{fig:Comparison2}
\end{figure}
\begin{figure}[t]
\begin{center}
\includegraphics[width=0.98\linewidth]{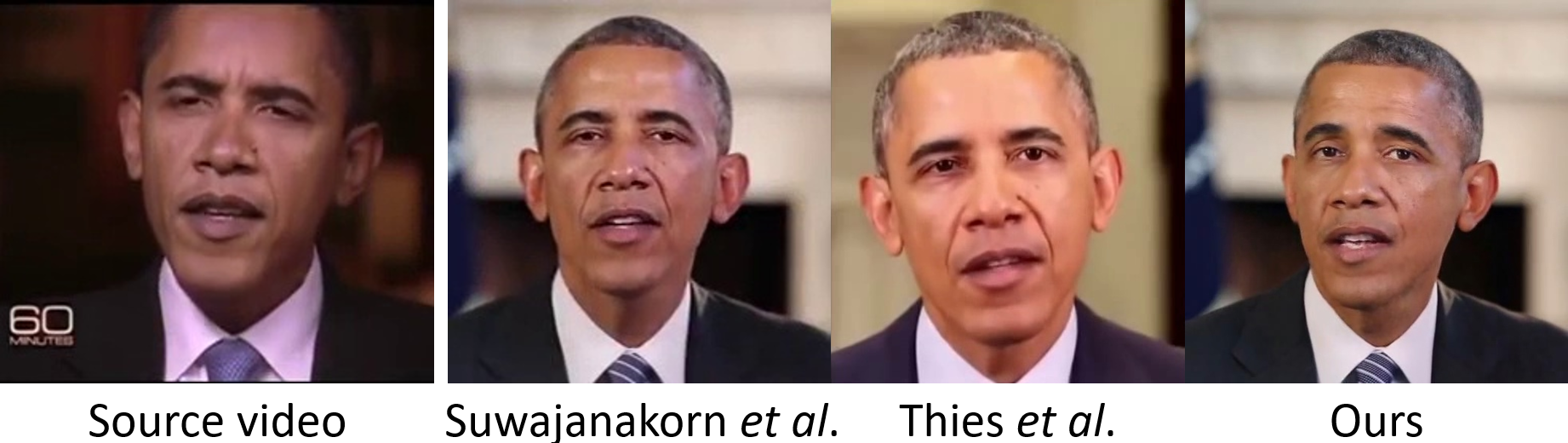}
\end{center}
\caption{Comparison to Suwajanakorn \textit{et al.}'s Synthesizing Obama~\cite{suwajanakorn2017synthesizing} and Thies \textit{et al.}'s Neural Voice Puppetry~\cite{thies2020neural}.}
\label{fig:Comparison3}
\end{figure}

\subsection{Comparison with State-of-the-Arts}
\subsubsection{Qualitative Comparison}
As shown in~\figref{fig:comparison1}, we first compare our results with four state-of-the-art audio-driven talking face video generation methods: ATVG~\cite{chen2019hierarchical}, DAVS~\cite{zhou2019talking}, Zhou~\etal~\cite{zhou2020makelttalk} and Yi~\etal~\cite{yi2020audio}. The ATVG and DAVS are 2D-based methods that take an audio sequence and a target image as input. The head pose and eye blink in their generated videos are fully static, which is a contradiction to the human sense.
Zhou~\etal~\cite{zhou2020makelttalk} uses face landmarks as an intermediate step to generate talking face videos. However, using landmark positions to represent them cannot fully capture head pose dynamics.
By using 3D face model, Yi~\etal~\cite{yi2020audio} generates photo-realistic talking videos. However, its generated head pose shows subtle movements, as can be seen from its tracemaps in~\figref{fig:comparison1}, and the eye blinks are completely static.
In contrast, with collaborative learning of explicit and implicit attributes, our method generates realistic talking face video with personalized head movements and realistic eye blinking. We also compare our method with 2D GAN-based Vougioukas \textit{et al.}'s~\cite{vougioukas2019realistic} and Chen \textit{et al.}'s~\cite{chen2020talking} methods in~\figref{fig:Comparison2}. The comparisons are conducted on the same characters, and our results are of higher visual quality than all other methods.

We further compare our method with the audio-driven facial reenactment methods~\cite{suwajanakorn2017synthesizing,thies2020neural}, which first generate the lip area that is in sync with the input audio, and compose it to an original video. We show qualitative results based on the same character - Barack Obama, and facial reenactment methods can generate photo-realistic talking videos in~\figref{fig:Comparison3}. However, their generated implicit attributes (\eg, head pose and eye blinks) are exactly from the original video, which means that the length of the generated video is limited by the reference video, otherwise, a special video connection technology must be used.

\begin{table*}
\centering{
\caption{Quantitative comparisons of state-of-the-art models and our model. Better values are highlighted in bold.}
\label{table:comparison}
\renewcommand{\arraystretch}{0.7}
\setlength{\tabcolsep}{1.8mm}
\begin{tabular}{ l | c | c | c c | c c | c  c}
\toprule
\multirow{2}{*}{Method} & \multirow{2}{*}{LMD} & \multirow{2}{*}{CPBD} &\multirow{2}{*}{AV offset} & \multirow{2}{*}{AV confidence}  & \multirow{2}{*}{blinks/s}  & \multirow{2}{*}{blink dur. (s)} & \multicolumn{2}{c}{Personalization} \\
\cline{8-9}
 &  & &  & &  &  & blinks & head pose \\
\midrule
  ATVG~\cite{chen2019hierarchical} & 5.31 & 0.119 & -1 & 4.048 & N/A & N/A  &  N/A& N/A\\
 DAVS~\cite{zhou2019talking} & 4.54 & 0.144 & -3 & 2.796 & N/A  & N/A  &  N/A& N/A\\
Zhou~\cite{zhou2020makelttalk} & 4.97 & 0.271  & -2 & 5.086 & 0.42  & 0.21  & 0.40 & 0.52\\
 Yi~\cite{yi2020audio} & 3.82 & 0.291 & -2 & 4.060 & N/A  &N/A  & N/A  & 0.30\\
 Ours & \textbf{3.57} & \textbf{0.314}  & -2 & \textbf{5.216} & 0.47 & 0.26 & \textbf{0.73} & \textbf{0.85}\\

\bottomrule
\end{tabular}
}
\end{table*}

\subsubsection{Quantitative Evaluation}
\textbf{Landmark distance metric:} 
We apply the Landmark Distance (LMD) proposed by Chen~\etal~\cite{chen2018lip} for accuracy evaluation of lip movement.

\noindent \textbf{Sharpness metric:} 
The frame sharpness is evaluated by the cumulative probability blur detection (CPBD).

\noindent \textbf{Lip-sync metric:} 
We evaluate the synchronization of lip motion with input audio by SyncNet~\cite{chung2016out}, which calculates the Audio-Visual (AV) Offset and Confidence scores to determine the lip-sync error.

\noindent\textbf{Eye blink metric:} 
The average human eye blinking rate is 0.28-0.45 blinks/s and average inter-blink duration is 0.41s~\cite{sinha2020identity}. These reference values will vary for different people and speaking scenarios.

\noindent\textbf{Personalization metric:}
One high-quality synthesizing should be able to generate personalized characteristics for different identities. To evaluate this personalization capability, we train a typical $N$-way pose classification network (see supplementary for more information) of $N$ identities by matching their input head poses or eye blinks.

As shown in~\tabref{table:comparison}, it can be found our model can generate personalized attributes and surpasses most existing methods~\cite{chen2019hierarchical,zhou2019talking,zhou2020makelttalk,yi2020audio}, which verifies the effectiveness of our collaborative learning network.

\begin{figure}[t]
\begin{center}
\includegraphics[width=0.98\linewidth]{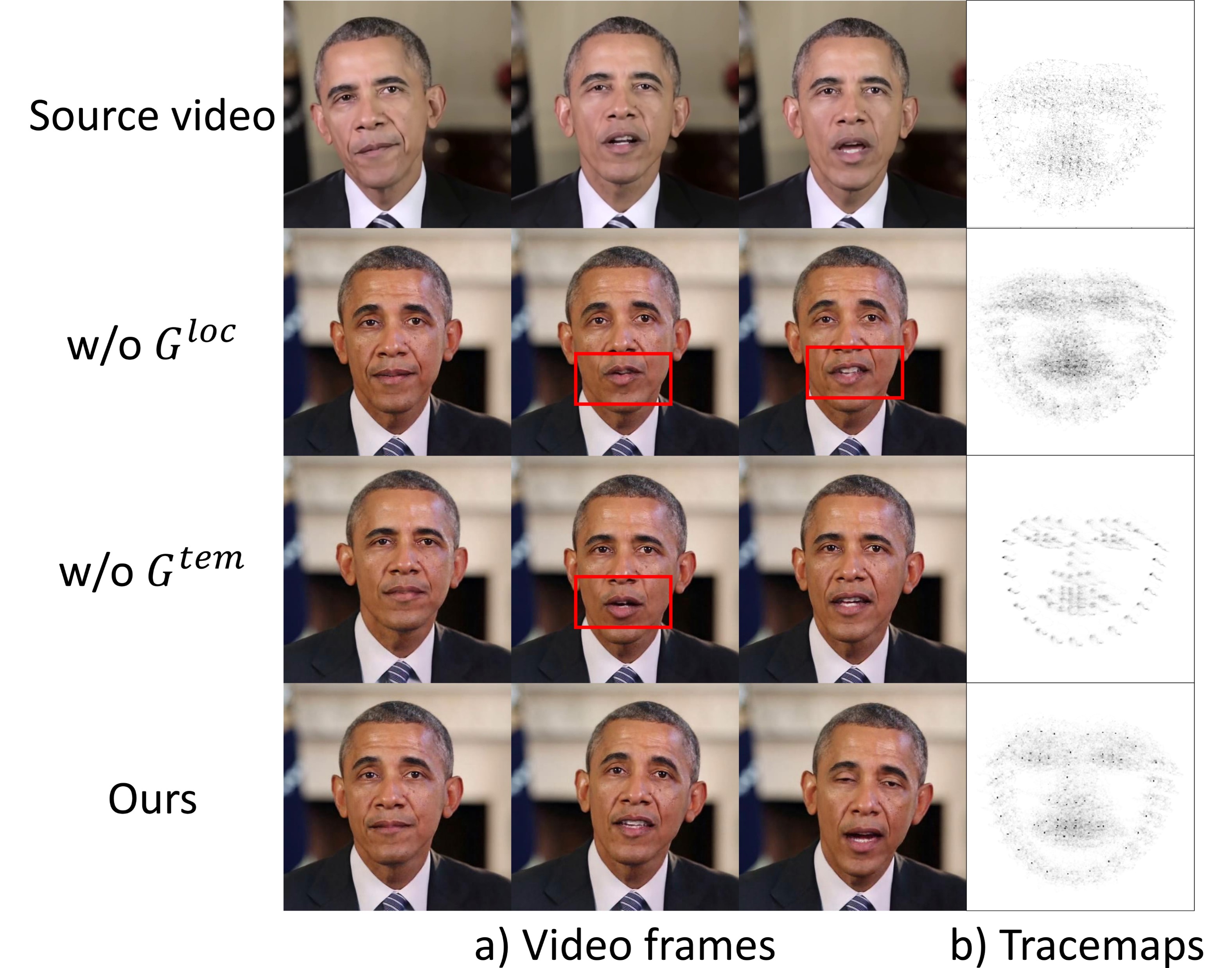}
\end{center}
\caption{Ablation study for $\br{G}^{tem}$ and $\br{G}^{loc}$.}
\label{fig:ablation1}
\end{figure}

\begin{figure}[t]
\begin{center}
\includegraphics[width=1.0\linewidth]{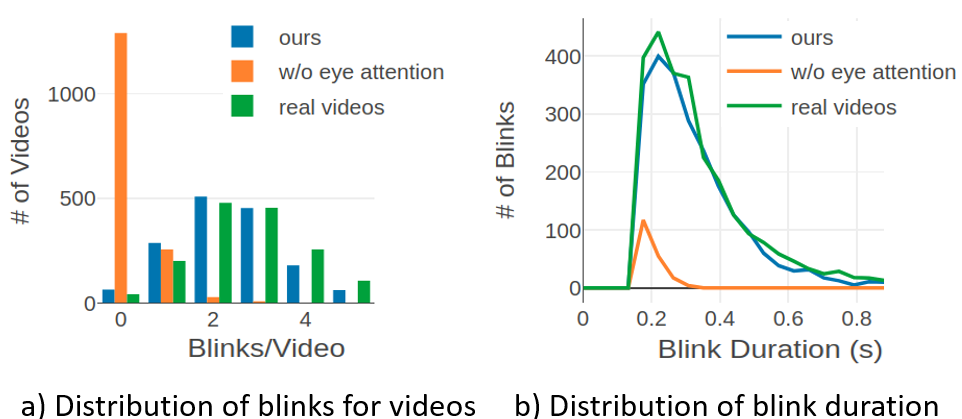}
\end{center}
\caption{Ablation study for part-sensitive encoding maps.}
\label{fig:ablation2}
\end{figure}

\subsection{Ablation Studies}
In our FACIAL-GAN module, we generate temporal correlation feature $\br{z}$ and local phonetic feature $\br{c}$, and then use decoders to convert those two features into facial attributes including expression, head pose, and eye blinking AU. Here we evaluate the importance of these two features. As shown in~\figref{fig:ablation1}, the generated videos result in -3 / 4.309 (AV offset / confidence) in second row (w/o $\br{G}^{loc}$), and -2 / 4.051 in third row (w/o $\br{G}^{tem}$), and -2 / 5.127 for our combined method. In addition, from the tracemaps we can see the head motion is more static without the $\br{G}^{tem}$ network.

We also evaluate our part-sensitive encoding module. For video frames without eye attentions, the blinking frequencies of generated videos are extremely low and unnatural. We sample 1,569 video clips from our testing dataset and each clip is about 4 seconds. Then we calculate the distribution of eye blinks for video clips and the distribution of blink duration time with and without eye attention maps in~\figref{fig:ablation2}. The blinking frequency and duration from the results of our method are similar to that of the real videos.

\begin{table}[t]
\begin{center}
\renewcommand{\arraystretch}{0.6}
\setlength{\tabcolsep}{1.8mm}
\caption{User study analyses of our model with state of the arts.}
\label{tab:retHumanStudy} 
\begin{tabular}{c |l|cccc}
\toprule
 \multirow{2}{*}{\begin{tabular}[c]{@{}c@{}}Implicit\\ Att.\end{tabular}}& \multirow{2}{*}{Method} & \multirow{2}{*}{Image} & \multirow{2}{*}{Lip} & \multirow{2}{*}{Pose} & \multirow{2}{*}{Blink}\\
 & & & & & \\

\midrule
\multirow{2}{*}{\xmark} & ATVG~\cite{chen2019hierarchical}  &-1.1 & 0.2 & -1.6 & -1.3 \\
 & DAVS~\cite{zhou2019talking}  &-1.2 & -1.7 & -1.7 & -1.6 \\
\midrule
\multirow{5}{*}{\cmark} & Zhou~\cite{zhou2020makelttalk} & 0.8 & 0.9 & 0.6 & 1.0 \\
 & Yi~\cite{yi2020audio} & \textbf{1.6} & 0.8 & 0.9 & 0.2 \\
 & Vougioukas~\cite{vougioukas2019realistic} & -0.6	& 1.1 & -1.3 & -1.4 \\
 & Chen~\cite{chen2020talking}  & -1.2 & 0.2 & 0.3 & -0.3 \\

 & Ours & \textbf{1.6} & \textbf{1.2} & \textbf{1.7} & \textbf{1.4} \\
\midrule
 & Real video & 1.9 & 2.0 & 1.9 & 2.0 \\
\bottomrule
\end{tabular}
\end{center}
\end{table}

\subsection{User Studies}
We conduct user studies to compare the generated results from the human perspective. 20 volunteers participate in the study to evaluate the video quality based on four criteria: 1) photo-realistic image quality, 2) audio-lip synchronization, 3) natural head motion, 4) realistic eye blinks.
Participants are required to evaluate each video 4 times based on the evaluation criteria. The evaluation scores include: -2 (very bad), -1 (bad), 0 (normal), 1 (good), 2 (very good). Every participant learns 3 examples first, and then evaluates 18 videos of either real video or synthesized from face generation methods. We calculate the average value of evaluated results for each method.
Participants’ evaluation results are summarized in Table~\ref{tab:retHumanStudy}, which indicates that our method is better than 
state-of-the-art methods.

\section{Discussion and Future Work}
In this work, we focus on implicit attribute learning targeting for natural head poses and eye blinks. It should be noted that human talking videos still have other implicit attributes,~\eg, gaze motion, body and hand gestures, microexpressions, etc., which are guided by other dimensions of information and may require specific designs of other network components. We hope our FACIAL framework could be a stepping stone towards the future exploration of implicit attribute learning along with those directions.

\section*{Acknowledgments}
This research is partially supported by National Science Foundation (2007661) and research gifts from Samsung Research America. Zeng was supported by NSFC (No.62072382), Fundamental Research Funds for the Central Universities, China (No.20720190003). The opinions expressed are solely those of the authors, and do not necessarily represent those of the funding agencies.

{\small
\bibliographystyle{ieee_fullname}
\bibliography{egbib}

\begin{thebibliography}{10}\itemsep=-1pt

\bibitem{agarwal2019protecting}
Shruti Agarwal, Hany Farid, Yuming Gu, Mingming He, Koki Nagano, and Hao Li.
\newblock Protecting world leaders against deep fakes.
\newblock In {\em Proceedings of the IEEE Conference on Computer Vision and
  Pattern Recognition Workshops}, pages 38--45, 2019.

\bibitem{amos2016openface}
Brandon Amos, Bartosz Ludwiczuk, and Mahadev Satyanarayanan.
\newblock Openface: A general-purpose face recognition library with mobile
  applications.
\newblock Technical report, 2016.

\bibitem{blanz1999morphable}
Volker Blanz, Thomas Vetter, et~al.
\newblock A morphable model for the synthesis of 3d faces.
\newblock In {\em Siggraph}, volume~99, pages 187--194, 1999.

\bibitem{bregler1997video}
Christoph Bregler, Michele Covell, and Malcolm Slaney.
\newblock Video rewrite: Driving visual speech with audio.
\newblock In {\em Proceedings of the 24th Annual Conference on Computer
  Graphics and Interactive Techniques}, pages 353--360, 1997.

\bibitem{cao2013facewarehouse}
Chen Cao, Yanlin Weng, Shun Zhou, Yiying Tong, and Kun Zhou.
\newblock Facewarehouse: A 3d facial expression database for visual computing.
\newblock {\em IEEE Transactions on Visualization and Computer Graphics},
  20(3):413--425, 2013.

\bibitem{chan2019everybody}
Caroline Chan, Shiry Ginosar, Tinghui Zhou, and Alexei~A Efros.
\newblock Everybody dance now.
\newblock In {\em IEEE International Conference on Computer Vision (ICCV)},
  pages 5933--5942, 2019.

\bibitem{chen2020talking}
Lele Chen, Guofeng Cui, Celong Liu, Zhong Li, Ziyi Kou, Yi Xu, and Chenliang
  Xu.
\newblock Talking-head generation with rhythmic head motion.
\newblock In {\em European Conference on Computer Vision (ECCV)}, pages 35--51,
  2020.

\bibitem{chen2018lip}
Lele Chen, Zhiheng Li, Ross~K Maddox, Zhiyao Duan, and Chenliang Xu.
\newblock Lip movements generation at a glance.
\newblock In {\em European Conference on Computer Vision (ECCV)}, pages
  520--535, 2018.

\bibitem{chen2019hierarchical}
Lele Chen, Ross~K Maddox, Zhiyao Duan, and Chenliang Xu.
\newblock Hierarchical cross-modal talking face generation with dynamic
  pixel-wise loss.
\newblock In {\em IEEE Conference on Computer Vision and Pattern Recognition
  (CVPR)}, pages 7832--7841, 2019.

\bibitem{chung2017b}
Joon~Son Chung, Amir Jamaludin, and Andrew Zisserman.
\newblock You said that?
\newblock In {\em British Machine Vision Conference (BMVC)}, 2017.

\bibitem{chung2016lip}
Joon~Son Chung and Andrew Zisserman.
\newblock Lip reading in the wild.
\newblock In {\em Asian Conference on Computer Vision (ACCV)}, pages 87--103,
  2016.

\bibitem{chung2016out}
Joon~Son Chung and Andrew Zisserman.
\newblock Out of time: automated lip sync in the wild.
\newblock In {\em Asian Conference on Computer Vision (ACCV)}, pages 251--263,
  2016.

\bibitem{cooke2006audio}
Martin Cooke, Jon Barker, Stuart Cunningham, and Xu Shao.
\newblock An audio-visual corpus for speech perception and automatic speech
  recognition.
\newblock {\em The Journal of the Acoustical Society of America},
  120(5):2421--2424, 2006.

\bibitem{deng2019accurate}
Yu Deng, Jiaolong Yang, Sicheng Xu, Dong Chen, Yunde Jia, and Xin Tong.
\newblock Accurate 3d face reconstruction with weakly-supervised learning: From
  single image to image set.
\newblock In {\em Proceedings of the IEEE Conference on Computer Vision and
  Pattern Recognition Workshops}, pages 0--0, 2019.

\bibitem{ekman1978manual}
Paul Ekman and Wallace~V Friesen.
\newblock {\em Manual for the facial action coding system}.
\newblock 1978.

\bibitem{ezzat2002trainable}
Tony Ezzat, Gadi Geiger, and Tomaso Poggio.
\newblock Trainable videorealistic speech animation.
\newblock {\em ACM Transactions on Graphics (TOG)}, 21(3):388--398, 2002.

\bibitem{hannun2014deep}
Awni Hannun, Carl Case, Jared Casper, Bryan Catanzaro, Greg Diamos, Erich
  Elsen, Ryan Prenger, Sanjeev Satheesh, Shubho Sengupta, Adam Coates, et~al.
\newblock Deep speech: Scaling up end-to-end speech recognition.
\newblock {\em arXiv preprint arXiv:1412.5567}, 2014.

\bibitem{johnson2016perceptual}
Justin Johnson, Alexandre Alahi, and Li Fei-Fei.
\newblock Perceptual losses for real-time style transfer and super-resolution.
\newblock In {\em European Conference on Computer Vision (ECCV)}, pages
  694--711. Springer, 2016.

\bibitem{kim2019neural}
Hyeongwoo Kim, Mohamed Elgharib, Michael Zollh{\"o}fer, Hans-Peter Seidel,
  Thabo Beeler, Christian Richardt, and Christian Theobalt.
\newblock Neural style-preserving visual dubbing.
\newblock {\em ACM Transactions on Graphics (TOG)}, 38(6):1--13, 2019.

\bibitem{kim2018deep}
Hyeongwoo Kim, Pablo Garrido, Ayush Tewari, Weipeng Xu, Justus Thies, Matthias
  Niessner, Patrick P{\'e}rez, Christian Richardt, Michael Zollh{\"o}fer, and
  Christian Theobalt.
\newblock Deep video portraits.
\newblock {\em ACM Transactions on Graphics (TOG)}, 37(4):1--14, 2018.

\bibitem{kingma2014adam}
Diederik~P Kingma and Jimmy Ba.
\newblock Adam: A method for stochastic optimization.
\newblock In {\em International Conference on Learning Representations (ICLR)},
  2015.

\bibitem{paysan20093d}
Pascal Paysan, Reinhard Knothe, Brian Amberg, Sami Romdhani, and Thomas Vetter.
\newblock A 3d face model for pose and illumination invariant face recognition.
\newblock In {\em 2009 Sixth IEEE International Conference on Advanced Video
  and Signal Based Surveillance}, pages 296--301, 2009.

\bibitem{prajwal2020lip}
KR Prajwal, Rudrabha Mukhopadhyay, Vinay~P Namboodiri, and CV Jawahar.
\newblock A lip sync expert is all you need for speech to lip generation in the
  wild.
\newblock In {\em Proceedings of the 28th ACM International Conference on
  Multimedia}, pages 484--492, 2020.

\bibitem{pumarola2018ganimation}
Albert Pumarola, Antonio Agudo, Aleix~M Martinez, Alberto Sanfeliu, and
  Francesc Moreno-Noguer.
\newblock Ganimation: Anatomically-aware facial animation from a single image.
\newblock In {\em European Conference on Computer Vision (ECCV)}, pages
  818--833, 2018.

\bibitem{ramamoorthi2001efficient}
Ravi Ramamoorthi and Pat Hanrahan.
\newblock An efficient representation for irradiance environment maps.
\newblock In {\em Proceedings of the 28th Annual Conference on Computer
  Graphics and Interactive Techniques}, pages 497--500, 2001.

\bibitem{sinha2020identity}
Sanjana Sinha, Sandika Biswas, and Brojeshwar Bhowmick.
\newblock Identity-preserving realistic talking face generation.
\newblock In {\em 2020 International Joint Conference on Neural Networks,
  (IJCNN)}, 2020.

\bibitem{song2018talking}
Yang Song, Jingwen Zhu, Dawei Li, Andy Wang, and Hairong Qi.
\newblock Talking face generation by conditional recurrent adversarial network.
\newblock In {\em Proceedings of the Twenty-Eighth International Joint
  Conference on Artificial Intelligence (IJCAI)}, pages 919--925, 2019.

\bibitem{suwajanakorn2017synthesizing}
Supasorn Suwajanakorn, Steven~M Seitz, and Ira Kemelmacher-Shlizerman.
\newblock Synthesizing obama: learning lip sync from audio.
\newblock {\em ACM Transactions on Graphics (TOG)}, 36(4):1--13, 2017.

\bibitem{thies2020neural}
Justus Thies, Mohamed Elgharib, Ayush Tewari, Christian Theobalt, and Matthias
  Nie{\ss}ner.
\newblock Neural voice puppetry: Audio-driven facial reenactment.
\newblock In {\em European Conference on Computer Vision (ECCV)}, pages
  716--731. Springer, 2020.

\bibitem{thies2016face2face}
Justus Thies, Michael Zollhofer, Marc Stamminger, Christian Theobalt, and
  Matthias Nie{\ss}ner.
\newblock Face2face: Real-time face capture and reenactment of rgb videos.
\newblock In {\em IEEE Conference on Computer Vision and Pattern Recognition
  (CVPR)}, pages 2387--2395, 2016.

\bibitem{vougioukas2019end}
Konstantinos Vougioukas, Stavros Petridis, and Maja Pantic.
\newblock End-to-end speech-driven realistic facial animation with temporal
  gans.
\newblock In {\em Proceedings of the IEEE Conference on Computer Vision and
  Pattern Recognition Workshops}, pages 37--40, 2019.

\bibitem{vougioukas2019realistic}
Konstantinos Vougioukas, Stavros Petridis, and Maja Pantic.
\newblock Realistic speech-driven facial animation with gans.
\newblock {\em International Journal of Computer Vision}, pages 1--16, 2019.

\bibitem{wang2018high}
Ting-Chun Wang, Ming-Yu Liu, Jun-Yan Zhu, Andrew Tao, Jan Kautz, and Bryan
  Catanzaro.
\newblock High-resolution image synthesis and semantic manipulation with
  conditional gans.
\newblock In {\em IEEE Conference on Computer Vision and Pattern Recognition
  (CVPR)}, pages 8798--8807, 2018.

\bibitem{wen2020photorealistic}
Xin Wen, Miao Wang, Christian Richardt, Ze-Yin Chen, and Shi-Min Hu.
\newblock Photorealistic audio-driven video portraits.
\newblock {\em IEEE Transactions on Visualization and Computer Graphics}, 2020.

\bibitem{wiles2018x2face}
Olivia Wiles, A Sophia~Koepke, and Andrew Zisserman.
\newblock X2face: A network for controlling face generation using images,
  audio, and pose codes.
\newblock In {\em European Conference on Computer Vision (ECCV)}, pages
  670--686, 2018.

\bibitem{yi2020audio}
Ran Yi, Zipeng Ye, Juyong Zhang, Hujun Bao, and Yong-Jin Liu.
\newblock Audio-driven talking face video generation with natural head pose.
\newblock {\em arXiv preprint arXiv:2002.10137}, 2020.

\bibitem{zakharov2019few}
Egor Zakharov, Aliaksandra Shysheya, Egor Burkov, and Victor Lempitsky.
\newblock Few-shot adversarial learning of realistic neural talking head
  models.
\newblock In {\em IEEE International Conference on Computer Vision (ICCV)},
  pages 9459--9468, 2019.

\bibitem{zhang2021face}
Chenxu Zhang, Saifeng Ni, Zhipeng Fan, Hongbo Li, Ming Zeng, Madhukar Budagavi,
  and Xiaohu Guo.
\newblock 3d talking face with personalized pose dynamics.
\newblock In {\em International conference on Computational Visual Media
  (CVM)}, 2021.

\bibitem{zhang2019one}
Yunxuan Zhang, Siwei Zhang, Yue He, Cheng Li, Chen~Change Loy, and Ziwei Liu.
\newblock One-shot face reenactment.
\newblock In {\em British Machine Vision Conference (BMVC)}, 2019.

\bibitem{zhou2019talking}
Hang Zhou, Yu Liu, Ziwei Liu, Ping Luo, and Xiaogang Wang.
\newblock Talking face generation by adversarially disentangled audio-visual
  representation.
\newblock In {\em AAAI Conference on Artificial Intelligence (AAAI)},
  volume~33, pages 9299--9306, 2019.

\bibitem{zhou2020makelttalk}
Yang Zhou, Xintong Han, Eli Shechtman, Jose Echevarria, Evangelos Kalogerakis,
  and Dingzeyu Li.
\newblock Makelttalk: speaker-aware talking-head animation.
\newblock {\em ACM Transactions on Graphics (TOG)}, 39(6):1--15, 2020.

\end{thebibliography}
}

\end{document}